\documentclass[]{article}

\usepackage{graphicx}
\usepackage{pstricks}
\usepackage{amssymb}
\usepackage{amsfonts}
\usepackage{amsmath}
\usepackage[amsmath,thmmarks]{ntheorem}
\usepackage{mathrsfs}

\newrgbcolor{todocolor}{.0 .4 .5}

\newrgbcolor{commentcolor}{.7 .0 .2}

\usepackage{multirow} 
\usepackage{rotating}

\usepackage{subfig}
\usepackage{color}
\usepackage{ogonek}
\usepackage{float}
\usepackage{url}

\usepackage{tikz}
\usetikzlibrary{positioning}
\usetikzlibrary{arrows}
\tikzset{
    >=stealth',
    blackarrow/.style={
           ->,
           thick,
           shorten <=2pt,
           shorten >=2pt,}
}

\usepackage{lineno}

\begin{document}




\title{On Classification with Bags, Groups and Sets\footnote{Preprint submitted to Pattern Recognition Letters}}
\author{Veronika Cheplygina, David M.J. Tax, Marco Loog}

\maketitle

\begin{abstract}

Many classification problems can be difficult to formulate directly in terms of the traditional supervised setting, where both training and test samples are individual feature vectors. There are cases in which samples are better described by sets of feature vectors, that labels are only available for sets rather than individual samples, or, if individual labels are available, that these are not independent. To better deal with such problems, several extensions of supervised learning have been proposed, where either training and/or test objects are sets of feature vectors. However, having been proposed rather independently of each other, their mutual similarities and differences have hitherto not been mapped out. In this work, we provide an overview of such learning scenarios, propose a taxonomy to illustrate the relationships between them, and discuss directions for further research in these areas. 
\end{abstract}


\section{Introduction}
In recent years, the field of pattern recognition has seen many problems that are difficult to formulate as regular supervised classification problems where (feature vector, label) pairs are available to train a classifier that, in turn, can predict labels for previously unseen feature vectors. A subset of these problems contains learning scenarios where (part of) the objects are represented by sets or \emph{bags} of feature vectors or \emph{instances}. Such learning scenarios include multiple instance learning~\cite{dietterich1997solving}, set classification~\cite{ning2008set}, group-based classification~\cite{samsudin2010nearest} and many others. In this paper we review these learning scenarios.  

There are several reasons why a bag representation might be chosen in a pattern recognition problem. The first reason is that a single feature vector is often too restrictive to describe an object. For example, in drug activity prediction, we are interested in classifying molecules as having the desired effect (active) or not. However, a molecule is not just a list of its elements: most molecules can fold into different shapes or conformations, which can influence the activity of that molecule. Furthermore, the number of stable shapes is different per molecule. A more logical choice is therefore to represent a molecule as a set of its conformations.

The second reason is that labels on the level of feature vectors are difficult, costly and/or time-consuming to obtain, but labels on a coarser level can be obtained more easily. For computer aided diagnosis applications, it can be very expensive for a radiologist to label individual pixels or voxels in an image as healthy or diseased, while it is more feasible to tag a full
image or some large image regions with a single label. Such coarsely labeled scans or regions can then be used for train a classifier and predict labels at the bag level, i.e., complete patient scans, or at the finer grained region or instance level, e.g., by labeling individual pixels or voxels.

Another reason to consider the labeling of bags of instances, instead of single feature vectors, is that there can be structure in the labels of the instances. For example, in face verification, where a video of a person is available, considering all the video frames jointly can provide more confident predictions than labeling each of the frames individually and combining the decisions. Similarly, neighboring objects in images, videos, sounds, time series and so forth are typically very correlated, and thus should not be classified independently.

\begin{figure*}%
\begin{center}
\includegraphics[width=0.7\textwidth]{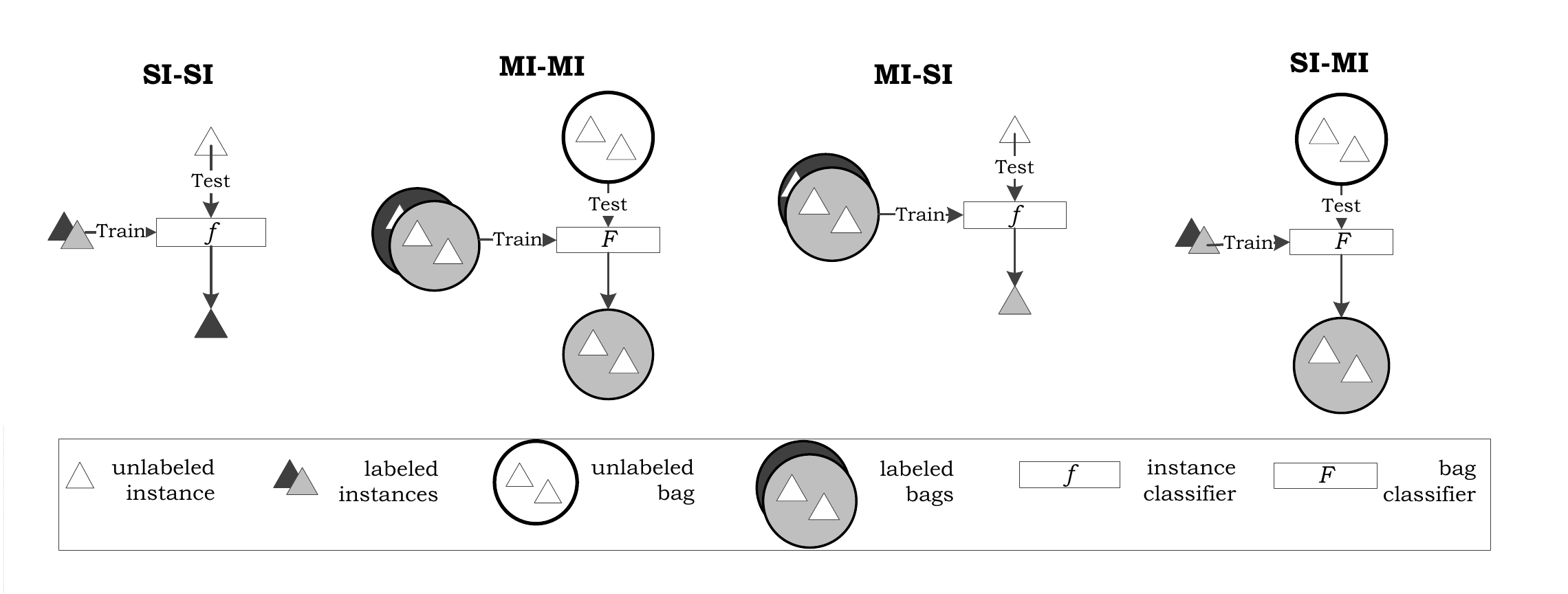}%
\caption{Supervised learning (SI-SI) and extensions. In the MI-MI scenario (Section~\ref{sec:bagsbags}), both training and test objects are bags. In the MI-SI scenario (Section~\ref{sec:bagsinst}), the training objects are bags and test objects are instances, while in the MI-SI scenario (Section~\ref{sec:instbags}), the training objects are instances and the test objects are bags.}%
\label{fig:chart}%
\end{center}
\end{figure*}

These examples have different goals and assumptions, and therefore may require different representations in the training and the test phase. All possibilities shown in Fig.~\ref{fig:chart} occur: both training and test objects can be single instances (SI) or bags, i.e. multiple instances (MI). Traditional supervised learning is in the SI-SI scenario, where both training and test objects are instances. Predicting molecule activity is in the MI-MI scenario, where both training and test objects are bags. Image classification problems can be found in the MI-MI scenario (training on images, testing on images) as well as the MI-SI scenario (training on images, testing on pixels or patches). The face verification problem is best represented by the SI-MI scenario (training on a single face, testing on a set of faces).


The success of a classifier in one application, such as molecule activity prediction, often motivates other researchers to use the same method in a different application, such as image classification. However, it is not necessarily the case that the assumptions of the first application still hold. For example, the assumptions on the relationships of bag and instance labels can be different for molecules and for images, which can lead to poor performances. On the other hand, it can also happen that the same type of problem occurs in two different applications, and that researchers in the respective fields approach the problem in different ways, without benefiting from each other's findings. We therefore believe that understanding the relationships between such learning scenarios is of importance to researchers in different fields.

With this work, our goal is to provide an overview of learning scenarios in which bags of instances play a role at any of the stages in the learning or classification process and to provide insight in their interconnections.  We have gathered papers that proposed novel learning scenarios, often combining synonyms of the word ``set'' with words such as ``classification'' or ``learning''. Our work is intended as a survey of learning problems, not of classifiers for a particular scenario, although we refer to existing surveys of this type whenever possible. Furthermore, we mainly focus on a single-label, binary classification scenario. Our focus is complimentary to the multi-label and/or multi-class setting and the problem formulations covered in this work can be extended to multi-label and multi-class. Examples can be found in~\cite{zhou2012multi,tsoumakas2007multi}. 

This paper begins with an overview of applications which motivate the bag representation in Section~\ref{sec:applications}, and the assumptions (such as on the relationship of instance and bag labels) associated with these applications. We then explain the categories of learning scenarios and the methodologies used to learn in such scenarios in Section~\ref{sec:methods}.  The paper concludes with a discussion in Section~\ref{sec:discussion}.


\section{Applications and Assumptions}\label{sec:applications}


\subsection{Molecule Activity Prediction}\label{sec:molecule}

In molecule activity prediction, the goal is to predict whether a previously unseen molecule has the desired activity, for example, whether a protein binds to another protein and thus influences a biological process. Often molecules have different conformations, or 3D shapes they can fold into, which influence their binding properties. Naturally, different molecules have different numbers of conformations. Therefore, one possibility is to represent molecules by the set of their conformations. For existing molecules, however, the information of which conformations are active, and which are not, is not available. A possible assumption in this case is that if at least one of the conformations is active, that the molecule can be regarded is active. This assumption is used in~\cite{dietterich1997solving} and~\cite{fu2012implementation} and entails that the instances have labels, and if at least one instance is positive, the bag is positive as well. 

Another possibility is to represent a molecule by a 3D cloud of atoms. Atom clouds with similar shapes are expected to display similar activity. Therefore, by aligning the clouds and comparing them directly, the function of previously unseen molecules can be predicted. This assumption is used in ~\cite{hoffmann2010new}. Here the instances (atoms) do not have labels, as it is not logical for an atom to be active or inactive, but certain combinations of instances do lead to different bag labels. In other words, most, or all instances contribute to the bag label.

\subsection{Image Classification}\label{sec:imclasf}

In one group of image classification applications, bags are images, and the instances are parts of the images, such as pixels, blobs or segments. Examples include natural scene classification~\cite{maron1998multiple,chen2006miles}, object recognition~\cite{andrews2002support,rahmani2005localized} or medical imaging~\cite{cheplygina2014classification,samsudin2010nearest,kalkan2012automated,kandemir2014computer}. Often the assumption is that not all parts of the image contribute to the image label. For example, in an image of a tiger, other surroundings can be present, or in a lung scan of a patient with a lung disease, healthy lung tissue can be present as well~\cite{cheplygina2014classification}. Each instance therefore has a label (positive, i.e. containing a tiger, or not) and a popular assumption, which we call the standard assumption, is that if at least one instance is positive, then the bag is also positive. The goal is to label novel images (i.e. bags). 

On the other hand, the standard assumption might not always be sufficient.  For example, if the instances are pixels, it might not be suitable to define pixels as belonging to the tiger concept. Perhaps a fraction of positive instances is more suitable. Or, for the beach concept, both instances containing sand and instances containing water might be needed, therefore asking for a conjunction of concepts. Relaxed assumptions to deal with such problems are described in~\cite{weidmann2003two}. 

Another assumption is that all instances in the bag share the same label. This assumption is used in ~\cite{samsudin2010nearest}, when classifying groups of cells as healthy or anomalous, with the added information that all cells in a group share the same label. Although training can be done using labeled cells, in the test phase, it might be advantageous to classify the cells jointly, rather than using a two-step approach where cells are classified first, and their decisions are combined. 

In general, the definition and generation of instances influences what is reasonable for the application at hand. Typically, the more knowledge is involved in generating the instances, the more assumptions could be applicable. Consider an application with photographs, where each photograph is labeled with the people in that photo. If we use as a face detector to generate candidate instances~\cite{guillaumin2010multiple}, it is reasonable to assume that each instance corresponds to one person in the photograph, as opposed to a situation where we randomly sample patches from the images. 


In another group of applications, instances are images, and bags are groups of images, such as videos. This setup is common for face recognition~\cite{ning2008set,wolf2003learning,kim2007discriminative,zhou2006sample}. For example, several images (such as from different cameras, or from different frames in a video) of the same person are available for training. Of course, the assumption here is that all the instances have the same label. Here the goals can be to label a single image, or a group of images.

\subsection{Image Annotation}\label{sec:imannot}

Image annotation is similar to image classification (Section~\ref{sec:imclasf}) in a sense that often, bags are images and instances are parts (pixels, blobs, segments) of those images. However, the goal here is different: instances, rather than bags, need to be labeled. For example, in~\cite{vezhnevets2010towards,muller2012multi} the goal is to label pixels or patches as belonging to the background, or one of the objects portrayed in the image. In~\cite{briggs2012rank}, the goal is to classify segments of spectrograms of bird song recordings as belonging to a particular bird species, but training on only spectrogram-level annotations. 

This goal can be achieved with supervised learning, by providing fully annotated training images, where each pixel or segment is labeled. However, providing such annotated images is costly, especially in medical imaging applications~\cite{kandemir2014computer} -- it is easier to only provide weakly annotated data (such as indicating an image, or an image frame where the foreground object is present). In this case, the assumption is that an image (indicated part of image) is positive if and only if it contains the object of interest, and negative if it does not.

Sometimes additional assumptions are used as well. For example, in \cite{kuck2005learning}, the bags are not only labeled with a category (such as ``tiger''), but also with a fraction of instances that contain tigers. More information is available about the label distribution of the output, therefore reducing the search space for the classifier. An even further constraint is that only one instance is allowed get a particular label, for example, when labeling a set of faces in a photograph with a set of names~\cite{kuncheva2010full}. Another common example is the assumption that spatially neighboring instances are correlated, and are therefore more likely to have the same label, such as regions of interest in medical images~\cite{vural2006batch}. 

Weakly annotated data is also a benefit in tracking~\cite{babenko2009visual}. Instead of providing instances (patches) of the tracked object to the learner, bags of patches (with several inexact locations of the tracked object) can be used to improve performance. However, the goal of the tracking algorithm is to again label patches (instances), not bags. 


\subsection{Document Classification}\label{sec:document}

A document, such as an article~\cite{andrews2002support,ray2005learning}, email discussion~\cite{zhou2009multi} or website~\cite{kriegel2004classification} can be represented as a collection of its parts, such as paragraphs or individual webpages, which are often described by bag-of-words histograms. In these applications, the goal is to assign a category to unlabeled documents. Again, different assumptions might be applicable here, which can be more or less appropriate depending on the types of documents and document categories in question. 

The assumption ``a positive bag has at least one positive instance'' seems applicable if we consider classifying biomedical articles as relevant or not for a particular gene ontology (GO) code. If at least one paragraph is relevant, then the whole article is considered relevant. In classifying more general-purpose documents, such as websites or email discussions, the situation might be different. For example, most social websites have a page describing the security settings, but it would be wrong to put these websites in the ``security'' category. An application where websites are classified is described in~\cite{kriegel2004classification}. Here a website is represented as a set of feature vectors, and no assumptions are made about the label relationships of the instance and bag labels.

In the above applications, the goal is to classify unlabeled bags. However, just as for images, for documents we can also be interested in instance labels, i.e., labeling individual emails~\cite{carvalho2005collective} or webpages~\cite{mcdowell2007cautious}. An assumption that is often used in such cases is that neighboring instances, such as webpages that link to each other, have correlated labels.  


\subsection{Others}\label{sec:other}

Other applications where the bag of instances representation has been used are detecting hard drive failures~\cite{murray2006machine}, detecting fraudulent financial accounts~\cite{juszczak2009behavioural}, music information retrieval~\cite{mandel2008multiple}, and spam filtering and advertising~\cite{quadrianto2009estimating}.

There are several reasons to motivate such representations. In some cases, only weak bag labels can be provided because it is not clear which instances correspond to these labels. For example, in hard drive failures, the bags are time series of different measurements of hard drives, and it is known for these hard drive whether a failure occurred or not. However, it is difficult to delineate the exact time frame that corresponds to the failure, and therefore multiple frames (instances) are used instead. 

In some cases, bag labels can be provided along with percentages of instance labels. For example, in spam filtering, it is possible to estimate proportions of spam/normal for a particular user, which helps to classify individual emails (instances) later on. In advertising, it can be estimated which proportion of customers would buy a product only on discount, and which proportion would buy a product in any circumstances. During an advertising campaign, these proportions can help to predict which customers (instances) should receive a discount coupon (and therefore buy the product). 

A rather different case from all others is addressing privacy issues, an application where instance labels (information about individuals) might be available, but these should not be shared or stored. Instead, it could be less problematic to provide labels about entire groups of people, such as the collective income~\cite{musicant2007supervised}, or the fraction of the group with a particular label. Based on such information, the goal is to label instances, such as assessing individual customers applying for a loan.


\section{Methodologies}\label{sec:methods}


\subsection{Notation and Overview}\label{sec:notation}

Mathematically, an instance is represented by a single feature vector $\mathbf{x} \in \mathcal{X}$, where $\mathcal{X} = \mathbb{R}^d$ is a $d$-dimensional space, while a bag is represented by a set of $n_i$ feature vectors $B_i=\{\mathbf{x}_{ik}; k=1...n_i\} \in 2^{\mathcal{X}}$. We denote the set of possible classes $\mathcal{C}$, and the set of possible labels $\mathcal{Y}$. In the case where each object has only one class label which we focus on in this overview, $\mathcal{Y}=\mathcal{C}$, in a multi-label scenario $\mathcal{Y} = 2^{\mathcal{C}}$. When a test object is an instance, we are interested in finding an instance classifier $f: \mathcal{X} \rightarrow \mathcal{Y}$. When a test object is a bag, we are generally interested in finding a bag classifier $F: 2^{\mathcal{X}} \rightarrow \mathcal{Y}$, or, in some special cases, $F: 2^{\mathcal{X}} \rightarrow 2^{\mathcal{Y}}$.  





We categorize the learning scenarios by the following characteristics: 

\begin{itemize}
\item \textbf{Type of training data} provided to the classifier: labeled instances, or labeled bags. In the case a bag is provided, usually the labels for the individual instances are not available. 
\item \textbf{Type of test data} classified by the trained classifier: instances or bags. In most cases this determines how evaluation is done: on instance level or on bag level. 

\item \textbf{Assumptions on labels}. Different applications have different assumptions of how the labels of the instances and the labels of the bags are related: for example, an assumption could be that all instances in a bag have the same label. These assumptions play an important role in how the learning algorithms are developed. 
\end{itemize}

These characteristics lead us to the categories in the leftmost column of Table~\ref{tab:settings}. In the following subsections, which are organized by the first two characteristics (types of training and test data), we explain each category, the corresponding learning scenarios and assumptions, the equivalence of different terms in literature, or why the category is empty. 

\begin{table*}
\centering
\caption{Learning scenarios: type of training and test data, assumptions on instance/bag labels, and main references.}
\begin{tabular}{l l l l p{5cm}}

Section & Train & Test & Assumptions & Main references \\
\hline

\ref{sec:instinst}. SI-SI& Instances & Instances & Weak & Supervised learning \\

& Instances & Instances	 & Strong & Batch classification\cite{vural2006batch}, Collective classification\cite{sen2008collective,carvalho2005collective,mcdowell2007cautious} \\

\hline

\ref{sec:bagsbags}. MI-MI  & Bags & Bags & Weak & Sets of feature vectors~\cite{kondor2003kernel,jebara2003images,kriegel2004classification} \\
  & Bags			& Bags 	 & Strong &  Multiple instance learning ~\cite{dietterich1997solving,maron1998framework} \\

\hline

\ref{sec:bagsinst}. MI-SI & Bags & Instances & Weak & - \\
& Bags & Instances & Strong &  Multiple instance learning ~\cite{muller2012multi,vezhnevets2010towards}, Aggregate output learning ~\cite{musicant2007supervised}, Learning with label proportions ~\cite{quadrianto2009estimating} \\

\hline
\ref{sec:instbags}. SI-MI & Instances & Bags & Weak & - \\

& Instances & Bags & Strong & Group-based classification~\cite{samsudin2010nearest}, Set classification~\cite{ning2008set}, Full-class set classification~\cite{kuncheva2010full} \\
 
\end{tabular}
\label{tab:settings}
\end{table*}

\subsection{SI-SI: Train on Instances, Test on Instances}\label{sec:instinst}

The first category of Table~\ref{tab:settings} contains traditional \textbf{supervised learning} where both training and test objects are assumed to be independently generated from some underlying class distributions. We assume that the reader is familiar with supervised learning. For a general introduction, refer to~\cite{jain2000statistical}. With the assumption of independently drawn train and test instances, the best possible approach is to train an instance classifier $f$ and classify each feature vector individually. However, in some situations data is not independently generated, and we can make more assumptions about the correlations in the data, and use these assumptions to improve the performance. 


    

    
    
   

The classical, rather general way to model dependencies between observations is through Markov random fields~\cite{kindermann1980markov} (MRFs) and the related, currently more popular conditional random fields (CRFs) ~\cite{lafferty2001conditional}. CRFs are originally described in the setting of labeling sequences, such as assigning part-of-speech tags to words a sentence, although other graph structures can also be defined. The goal is a word classifier $f$ where the output space $\mathcal{Y}$ is the set of all part-of-speech tags. To account for dependencies between parts-of-speech, the classifier that is used is a bag classifier $F$, trained on labeled sentences, rather than labeled words. The output space of this classifier is $2^{\mathcal{Y}}$, i.e. all possible combinations of parts-of-speech. Labels in this space, of course, can be ``dissected'', to provide instance labels $y \in \mathcal{Y}$ for the sentence classifier we were originally interested in. Performance is evaluated on instance level. Learning is therefore achieved by converting the SI-SI learning task into a MI-MI learning task. 

In \textbf{batch classification}~\cite{vural2006batch}, labeled instances are available for training and the goal is to label instances, therefore the task is in the SI-SI category. However, the authors observe that in their application (labeling ROIs in medical images) correlations exist between the instances in a bag, therefore it is more advantageous to label bags of instances instead. The correlations are provided in a covariance matrix of the instances. An instance classifier $f$ with bag-level constraints (derived from the correlations) is trained first. In the test phase, an instance $\mathbf{x}$ is classified by a weighted average of instances $\mathbf{x}_i$, correlated with $\mathbf{x}$. Although this is not done explicitly, we can also see this learning approach as a way to convert a SI-SI task to a MI-MI task.

In \textbf{collective classification}~\cite{sen2010collective,carvalho2005collective,mcdowell2007cautious}, the goal is to label instances, given that correlations exist between these instances. \cite{sen2010collective} distinguishes two types of approaches for this, which they call local and global. In one of the local approaches, instance classifiers are trained, although relational features, i.e., features encoding the labels of the correlated instances, are also used. In the test phase, after an initial prediction, the label of each test instance is updated based on the labeling of the other test instances. This, in turn, changes the relational features. The process is repeated iteratively. Thus, only instance classifiers $f$ are used, but the bag-level constraints are for the most part encoded in the feature representation, rather than in the learning algorithm. In one of the global approaches, MRFs are used to simultaneously predict the instance labels, therefore using a bag classifier $F$.

\subsection{MI-MI: Train on Bags, Test on Bags}\label{sec:bagsbags}

When both the training objects and test objects are bags, but no additional assumptions about the labels are present, the goal is \textbf{classification of sets of feature vectors}~\cite{kondor2003kernel} (not to be confused with set classification which is an unfortunate name for a different scenario, discussed in Section~\ref{sec:instbags}). As a result, the only possible strategy is to train the bag classifier $F$ by comparing bags directly. This is possible by defining distances or kernels on bags, or embedding the bags in a vector space. 

A well-known kernel for bags~\cite{gartner2002multi,hoffmann2010new} is the convolution kernel, in which all instances of one bag are compared to all instances of another bag: $K(B_i,B_j) =  \sum_{k} \sum_{l} k(\mathbf{x}_{ik}, \mathbf{x}_{jk})$ where $k$ is a kernel on feature vectors, such as a Gaussian kernel. This assumption that is implicitly made here is that all instances contribute to the bag label. A similar assumption is made in works which regard bags as samples from probability distributions, and define the kernel through a divergence~\cite{zhou2006sample,muandet2012learning}. For distances, the Hausdorff distance and its variants~\cite{kriegel2004classification,wang2000solving} also introduce certain assumptions. For example, the definition $d(B_i, B_j)  =  \min_{k}\min_{l}	d(\mathbf{x}_{ik}, \mathbf{x}_{jl})$ assumes that only the most similar instances contribute to the similarity between bags.


An alternative approach to learn $F$ is to define a single instance representation for each bag, therefore embedding the bags into a vector space. This can be done by summarizing instance statistics in each bag~\cite{gartner2002multi}, bag of words representations~\cite{vijayanarasimhan2008keywords} or representing a bag by its distances to the training data~\cite{cheplygina2014multiple}. Any standard supervised classifier can be used on this representation. In a sense, the problem has been converted to a SI-SI learning task. 

Another domain where both training and test objects are bags, but stronger assumptions are made is called \textbf{multiple instance learning} (MIL)~\cite{dietterich1997solving,maron1998framework}. In MIL, the objects are referred to as \emph{bags} of \emph{instances}. Originally, it was assumed that $\mathcal{Y} = \{-1, +1\}$, and that the bag labels are determined by the (hidden) labels of their instances: a bag is positive if and only if there is at least one positive instance inside the bag; a bag is negative if and only if all of its instances are negative. There are two main approaches to achieve the goal of classifying bags. Due to the assumption on the relationship of the bag and instance labels, earlier methods focused on first finding an instance classifier $f$, and then applying a combining rule $g$ to the instance outputs. To use the traditional assumption in MIL, $g$ is defined by the noisy OR function, as follows:

\begin{equation*}
F(B) = \begin{cases} +1 &\mbox{if } g(\{f(\mathbf{x}_k) \}_{k=1}^{n}) > 1 \\ 
-1 & \mbox{otherwise } \end{cases} 
\end{equation*}

\begin{equation}\label{eq:noisyrule}
g(\{f(\mathbf{x}_k) \}_{k=1}^{n})    = \frac{1 - \prod_{k=1}^n 1-f(\mathbf{x}_k)}{\prod_{k=1}^n 1-f(\mathbf{x}_k)}
\end{equation}

where $f(\mathbf{x}_k) = p(y_k=1|\mathbf{x}_k)$.


More relaxed formulations of the traditional assumption have also been proposed~\cite{weidmann2003two,foulds2010review}. For instance, for a bag to be positive, it needs to have a specific fraction of positive instances. With such alternative assumptions, it is still possible to find $f$ first and then apply an appropriate $g$ to determine the labels of the test bags. By assuming that all instances contribute to the bag lab independently, for instance, $g$ can be replaced by the product rule or other generalized rules~\cite{loog2004static} of combining instance posterior probabilities. 

Several MIL methods have moved away from using explicit assumptions on the relationships of instance and bag labels~\cite{foulds2010review}, and learn using assumptions on bags as a whole, therefore taking a detour to the ``set classification'' scenario above. In other words, such methods aim at finding $F$ directly rather than through a combination of $f$ and $g$. The approaches that can be applied here are the same as above, i.e. by defining distances, kernels, or by embedding the bags into a vector space. Most of the approaches used in practice implicitly assume that all instances contribute to the bag label. More extensive surveys of MIL assumptions and classifiers can be found in~\cite{zhou2004multi,foulds2010review,amores2013multiple}. 



%

\subsection{MI-SI: Train on Bags, Test on Instances}\label{sec:bagsinst}

This section is concerned with the case where training data is only labeled on bag-level, while instance-level labels are desired in the test phase. Note that this is not possible if no assumptions are made about the label transfer between instances and bags. This is why the ``MI-SI, weak assumptions'' category in Table~\ref{tab:settings} is empty (denoted by -). By making additional assumptions, however, something can be said about the instance-level labels of the test data. 

The standard assumption in \textbf{multiple instance learning} is one of the possibilities we can use to train the classifier using labeled bags, but provide instance-level labels for the test data. Although originally, the goal of MIL was to train a classifier $F$ and provide labels for bags, a side-effect of some algorithms (which define $F$ through a combination of instance-classifiers $f$) is that instance labels are predicted as well. The fact that only bag labels are required to produce instance labels means that less labels are required than in the usual supervised setting. 

The goals of classifying instances and classifying bags are not identical, and therefore, in many cases, the optimal bag classifier is not the optimal instance classifier and vice versa. An important reason in MIL for this is the standard assumption. If bag classification is done by combining instance predictions, such as in (\ref{eq:noisyrule}), false negative instances are going to have less effect on the bag performance than false positive instances. Consider a positive bag where a positive instance is misclassified as negative: if the bag has any other positive instances, or a negative instance that has been falsely classified as positive, the bag label will still be correct. However, for a negative bag the label changes as soon as a single instance is misclassified. Similar observations have been made in \cite{ray2005supervised} and in \cite{tragante2011instance}. A more general reason why the optimal instance and bag classifiers do not necessarily correspond, is unequal bag sizes. Misclassifying a bag with a few instances has less effect on the instance performance, than misclassifying a bag with many instances. The goals of the user (optimizing performance on instances) and the goal of the classifier (optimizing performance on bags) are therefore not matched, and instance labels in such cases should be used with caution.

At this point it is important to mention that learning with such weakly annotated data has links to \textbf{semi-supervised learning}~\cite{chapelle2006semi,zhu05survey} and \textbf{learning with only positive and unlabeled data}~\cite{elkan2008learning}. Both of these fields deal with weakly annotated data in a sense that some of it is annotated, and some of it is not. In multiple instance learning, all of the data (in the form of bags) is annotated, however, from the perspective of instances, these annotations are weak. Because the semi-supervised and positive-and-unlabeled scenarios do not deal with bags in either stage of the classification process, we do not further elaborate on them in this survey, however, further connections between these fields can be found in~\cite{zhou2007relation,li2013link}.



Other scenarios where only training objects are bags are \textbf{learning about individuals from group statistics}~\cite{kuck2005learning}, \textbf{aggregate output learning}~\cite{musicant2007supervised} and \textbf{learning with label proportions}~\cite{quadrianto2009estimating}, independent names for very related ideas. Here the bag labels are not just class labels, but proportions of class labels, $Y=\{y_i | i=1,\ldots,|\mathcal{C}|, y_i \in \mathbb{R}, \sum_i y_i = 1\}$. For instance, a bag can be labeled as ``75\% positive, 25\% negative''. These scenarios can be seen as a subset of multiple instance learning, where the fraction of positive instances (often called the witness rate) in the bags is already specified. An exact fraction is a stronger assumption than a non-zero fraction, therefore it should be easier to learn when the witness rate is given. For real-life MIL datasets, \cite{kuck2005learning} assumes that a positive bag has exactly $\frac{1}{n_i}$ positive instances. Other MIL methods take advantage of this by estimating a witness rate first, and then using this estimate to build instance classifiers~\cite{li2012multiple,gehler2007deterministic}. 


\subsection{SI-MI: Train on Instances, Test on Bags}\label{sec:instbags}

We now turn to the scenario where instance-level labels are available for training, but bag-level labels are needed in the test phase. If no assumptions are made about how the instance and bag labels are related, this is an impossible task, and the reason the category corresponding to SI-MI with few assumptions in Table~\ref{tab:settings} is empty. However, similarly to the SI-SI approaches with additional assumptions in Section~\ref{sec:instinst}, dependencies between the feature vectors inside a test bag can be exploited to improve the overall classification. The difference between the methodologies described here and in Section~\ref{sec:instinst} is that here, we are interested in labeling test bags and not instances. 

This situation occurs in \textbf{group-based classification}~\cite{samsudin2010nearest,brossi2012comparison} and \textbf{set classification}~\cite{ning2008set}, independently proposed names for the setting where test objects are sets of feature vectors from the same class. Note that this setting can be easily transferred to the MI-MI scenario, because if the instances in one bag have the same label, it is straightforward to create bags from instances and vice versa.  

In~\cite{samsudin2010nearest}, the classification of a test bag distance-based and is done by modifying the supervised versions of the nearest neighbor or the nearest mean classifiers. There are two broad approaches called the voting and the pooling scheme. In the voting scheme, each instance is labeled by a classifier $f$, such as the nearest neighbor, and the labels are combined with majority voting as $g$. In the pooling scheme, the distances are aggregated first, and only then converted to a label for the bag. The results show that the pooling scheme (i.e. a nearest neighbor classifier $F$ applied on the bag distances) produces better results. Similar results are obtained in~\cite{kalkan2012automated}, where classification of instances (patches in histopathological images) is done on two levels: instances and bags. Although some instance-level labels are available and an instance classifier can be built, considering the bag-level labels is still beneficial for performance.


Several approaches are studied in~\cite{ning2008set}. The most straightforward approach involves combining predictions of each instance in a bag during the test phase, i.e. defining $F$ as a combination $g$ of several instance classifiers. The best performing approach borrows from the MI-MI scenario, because in both the training and test phase, instance subsets are generated. Kernels are defined on these subsets, and the test bag is classified by combining the predictions of its subsets. 

\begin{figure}
\begin{center}
\includegraphics[width=0.7\columnwidth]{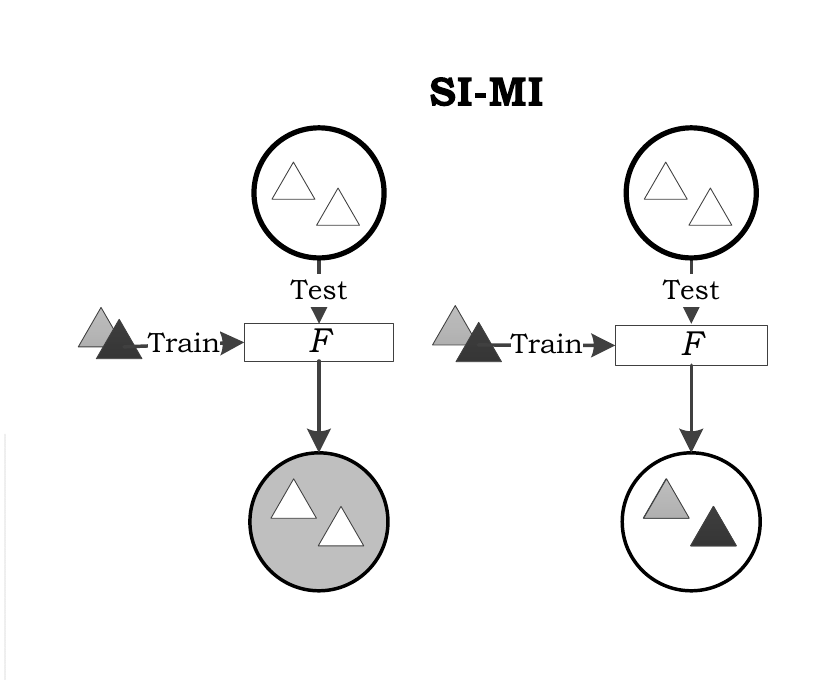}
\caption{Variants of the SI-MI scenario. The training objects are instances and the test objects are bags, although the bag can be labeled by a set of instance labels (right). In this case, the instance labels are decided jointly (as a bag super-label) by a bag classifier $F$, not by an instance classifier $f$.}
\label{fig:chart_inst_bags}
\end{center}
\end{figure}

The added information that all instances in a set share the same label is just one of the examples of a setting where the testing objects are bags. A reversed setting is \textbf{full-class set classification}~\cite{kuncheva2010full}. It has an additional constraint that each of the instances has a unique label, i.e. it is known beforehand which instance labels are present in the bag. Here the output of the bag classifier is not a single class label, but a super-label $Y \in \mathcal{I}$, where $\mathcal{I}$ is the set of permutations of the all class labels. Because $|\mathcal{I}| < |2^{\mathcal{C}}|$, \cite{kuncheva2010full} shows that a classifier $F$ that finds the instance labels jointly is guaranteed to perform better than concatenating the outputs of instance classifiers $f$. Note that although instance labels are obtained, the labels we are interested in (the super-labels) are bag labels, and the performance is evaluated on bag level: either all instances were labeled correctly, or not. We illustrate this with the diagrams in Fig.~\ref{fig:chart_inst_bags}. 

\section{Discussion}\label{sec:discussion}

Many classification problems deal with objects that are represented as sets of feature vectors, or so-called bags of instances. This popularity is not surprising, as there are several motivating reasons for choosing such a representation at one or more stages of the classification process. Firstly, a set of feature vectors provides greater representational power than a single feature vector, and it might not be logical to express multiple entities (such as several face images of one person) as a single entity. Secondly, often labels might be available only on bag level, and too costly to obtain on instance level, therefore using the bag of instances representation as a form of weak supervision. Lastly, it can be advantageous to consider bags as a whole rather than as independent instances, because of relationships of the instances in a single bag. 

We presented a taxonomy that illustrates the relationships of scenarios that deal with bags into four categories: SI-SI, MI-MI, MI-SI and SI-MI, according to whether single instances (SI) or multiple instances (MI) are available in the training and test phases of the learning scenarios. With this taxonomy, it becomes clear that the popularity of the bag representation also has dangers: several different learning scenarios are sometimes defined for the same problem (such as the SI-MI scenarios set classification and group-based classification), or several different problems are incorrectly grouped under the same learning scenario (such as the MI-MI and MI-SI scenarios for multiple instance learning). This may hinder research progress, because connections between existing learning scenarios are missed, or because erroneous connections, and therefore erroneous assumptions, are made. 

The algorithms used across the four categories are very diverse, as many supervised methods, such as the nearest neighbor classifier have been extended to work in these learning scenarios. An important observation across all these algorithms is that there are two main approaches: direct, where the training is done on the same type of input (SI or MI) as is originally available, and indirect, where training occurs via converting the problem to a different scenario, usually with additional assumptions. As canonical examples, consider a training set of labeled bags of unlabeled instances, and a test set of unlabeled bags (i.e. MI-MI category). A example of a direct approach is to define distance on bags and use a nearest neighbor classifier. An example of an indirect approach is to assume that all instances have the same label as the bag, train a instance classifier, and combine the instance labels in the test phase, i.e. solving the MI-MI problem via a SI-SI approach. 




While the proposed taxonomy allows for heterogeneity in training and test objects (i.e., SI-MI and MI-SI), it is limited because the training or test objects themselves are homogeneous. It would be interesting to investigate what happens in the case where in the training phase both labeled bags and labeled instances are available, such as in~\cite{kalkan2012automated}. As we already discussed in Section~\ref{sec:bagsinst}, the optimal bag classifier does not necessarily correspond with the optimal instance classifier. Therefore, deciding how to best use the available labels should depend on whether bags or instances are to be classified in the test phase. However, what if bags and instances can be expected in both the classification and test phases? A straightforward solution would be to train separate bag and instance classifiers, but when the bag and instance labels are related, an integrated classifier would perhaps be more suitable.

Another interesting observation is that the ``hybrid'' categories in the taxonomy (SI-MI and MI-SI) have attracted a lot of attention, and that the learning scenarios proposed here all need to rely on strong assumptions about the relationships of the instance and bag labels. One of the questions this raises is, what are the minimal assumptions needed to learn in such situations? Furthermore, the learning scenarios we reviewed do not exhaustively cover the types of constraints that could be present between the instance and bag labels. Learning scenarios that will be proposed in the future to fill some of these gaps, can now be easily placed in the context of the works described in this overview.

A development that would be very beneficial for this field is a collection of instance-labeled benchmark datasets where several scenarios can be adopted. This would enable not only comparisons of algorithms for a single scenario, as is often done in the literature, but the comparison of different learning scenarios, and thus, how suitable they are for the problem at hand.


\section*{Acknowledgments}

The authors would like to thank Brijnesh Jain, for his kind suggestions, leading us to improve the paper. The anonymous reviewers are kindly acknowledged for their critical comments and suggestions.

\bibliographystyle{abbrv}	%
\bibliography{publications_short}

\end{document}